\documentclass[10pt,twocolumn,letterpaper]{article}

\usepackage{cvpr}
\usepackage{times}
\usepackage{epsfig}
\usepackage{graphicx}
\usepackage{amsmath}
\usepackage{amssymb}

\usepackage{textcomp}
\usepackage{multirow}
\usepackage{subfigure}
\usepackage{graphicx}
\usepackage[small]{caption}
\usepackage{tabulary}
\usepackage{balance}

\newcommand{\bd}[1]{\textbf{#1}}
\newcommand{\app}{\raise.17ex\hbox{$\scriptstyle\sim$}}

\newcommand{\mrtwo}[1]{\multirow{2}{*}{#1}}

\usepackage[breaklinks=true,bookmarks=false]{hyperref}

\cvprfinalcopy 

\def\cvprPaperID{173} 

\begin{document}

\title{FocalMix: Semi-Supervised Learning for 3D Medical Image Detection}

\author{Dong Wang$^{1}$\thanks{Equal contribution.}
    ~~~~~Yuan Zhang$^{2}$\footnotemark[1]
    ~~~~~Kexin Zhang$^{2,3}$\thanks{Part of this work was done during internship at Yizhun Medical AI.}\footnotemark[2]
    ~~~~~Liwei Wang$^{1, 2}$\\
    $^1$Center for Data Science, Peking University\\
    $^2$Key Laboratory of Machine Perception, MOE, School of EECS, Peking University\\
    $^3$Yizhun Medical AI Co., Ltd\\
    {\tt\small \{wangdongcis,yuan.z,zhangkexin,wanglw\}@pku.edu.cn}
}

\maketitle
\thispagestyle{empty}

\begin{abstract}
Applying artificial intelligence techniques in medical imaging is one of the most promising areas in medicine. However, most of the recent success in this area highly relies on large amounts of carefully annotated data, whereas annotating medical images is a costly process. In this paper, we propose a novel method, called FocalMix, which, to the best of our knowledge, is the first to leverage recent advances in semi-supervised learning (SSL) for 3D medical image detection. We conducted extensive experiments on two widely used datasets for lung nodule detection, LUNA16 and NLST. Results show that our proposed SSL methods can achieve a substantial improvement of up to 17.3\% over state-of-the-art supervised learning approaches with 400 unlabeled CT scans. 

\end{abstract}

\section{Introduction}
Medical imaging plays an essential part in modern medical practice. One of the significant trends in this area is to exploit advanced techniques in deep learning (DL) and artificial intelligence (AI) to achieve automatic medical image analysis. 
Prior work has already demonstrated promising results in various specific tasks, such as skin cancer classification~\cite{esteva2017dermatologist}, retinal fundus image analysis~\cite{gulshan2016development}, with some preliminary real-world applications, e.g., \cite{chen2018augmented}. However, we argue that the success should be attributed to not only recent progress in deep learning techniques but also large volumes of carefully annotated data.

On the one hand, annotating medical images is an expensive and time-consuming process. This process requires experienced clinical experts to read examination reports, combine them with other test results, and sometimes consult with other experts. Furthermore, it is even more difficult to manually annotate such 3D images as CT and MRI with substantially more information. 
On the other hand, there are a large number of raw medical images stored in hospital information systems. The cost of retrieving them is negligible relative to the high expenses of human annotation.
Therefore, it becomes a necessary research question whether we can leverage these raw medical images with little annotation to improve the diagnostic accuracy of deep learning models.

Meanwhile, semi-supervised learning (SSL) has attracted a lot of research efforts in recent years. Most of the latest SSL methods generally add an auxiliary loss term defined on unlabeled data (e.g., consistency regularization term~\cite{sajjadi2016regularization}), or even linear interpolations of both labeled and unlabeled data (i.e., MixUp augmentation~\cite{DBLP:conf/iclr/ZhangCDL18}),
into the loss function for better generalization capacities and hence better performances on the test set. Some of them have achieved great success on image classification datasets such as CIFAR~\cite{cifar}, which fully demonstrates the potential value of utilizing unlabeled data. 

Applying recent advances of SSL to medical imaging problems seems to be a tempting approach. However, since people are more concerned with lesion detection tasks in medical imaging as opposed to the widely studied classification task in the existing SSL literature, many technical details remain unexplored. For instance, modern SSL frameworks generally require the loss function to be able to deal with soft labels (e.g., a smooth probability over classes), whereas most one-stage lesion detection models use the focal loss~\cite{lin2017focal}, which has no such natural extension. Also, state-of-the-art SSL methods use average ensembles to obtain pseudo-labels for unlabeled data. Nonetheless, it is hard to take the average over bounding boxes predicted by detection models. Last but not least, very few researches have touched on data augmentation for medical images, which, however, is almost an indispensable component for SSL approaches to achieve their recent success.

In this paper, we discuss a principled method, called FocalMix, for tailoring modern SSL frameworks to overcome the issues mentioned above.
First, we propose a generic generalization of the focal loss that allows the usage of \textit{soft-target} training labels with skewed distributions (analogous to class imbalance in discrete cases as encountered by most detection models) in Sect.~\ref{sect: soft focal}.
Then, practical designs are introduced to illustrate how we can extend essential components in an SSL framework for 3D medical image detection.
Specifically, we propose a \textit{target prediction} strategy that leverages anchor-level ensembles of augmented image patches by rotation and flipping (Sect.~\ref{sec:anchorprediction}).
Furthermore, the \textit{MixUp augmentation} is adapted for medical image detection tasks at both the \textit{image} level and \textit{object} level in light of unique characteristics of the medical image detection tasks (Sect.~\ref{sect: mixup for detection}). 
Throughout this paper, we mostly take a state-of-the-art SSL method, MixMatch~\cite{DBLP:journals/corr/abs-1905-02249}, as a running example to provide a more clear and approachable presentation. The proposed method can be transferred to other modern SSL frameworks (e.g., UDA~\cite{DBLP:journals/corr/abs-1904-12848}) with little effort.

Through extensive experiments on two widely-used datasets for pulmonary nodule detection on CT scans, we show that the proposed SSL method, FocalMix, can substantially outperform well-tuned state-of-the-art supervised learning approaches (Sect. \ref{sect: main results}). Ablation study further demonstrates the effectiveness of our proposed soft-target loss function, ensemble method for target prediction, and two levels of MixUp strategies (Sect. \ref{sect: ablation}). In addition, the results show that FocalMix can still boost the performance of supervised learning when there is a reasonably large annotated dataset already available (Sect. \ref{nlst_exp}). 

To conclude, the main contributions of this paper are:
\begin{itemize}
    \item We propose FocalMix, a novel semi-supervised learning framework for 3D medical image detection. 
    \item To the best of our knowledge, our work is the first to investigate the problem of semi-supervised learning for medical image detection.
    \item Through extensive experiments, we demonstrate that the proposed semi-supervised approach can significantly improve the performance of fully-supervised learning approaches.
\end{itemize}

\section{Background and Preliminaries}

\subsection{Object Detection in 3D Medical Images}

\label{sect: retinanet}

This paper mainly focuses on the problem of 3D medical image detection, which is an essential task in medical image analysis. In order to detect lesions of different scales, most works adopted anchor-based detectors, such as 3D variants of feature pyramid networks (FPN)~\cite{lin2017feature}. Meanwhile, the focal loss is widely used to overcome the extreme foreground-background class imbalance~\cite{lin2017focal}. This section provides a brief introduction to these methods.

\subsubsection{Anchor boxes}
\begin{figure}
\centering
\includegraphics[width=2.5in]{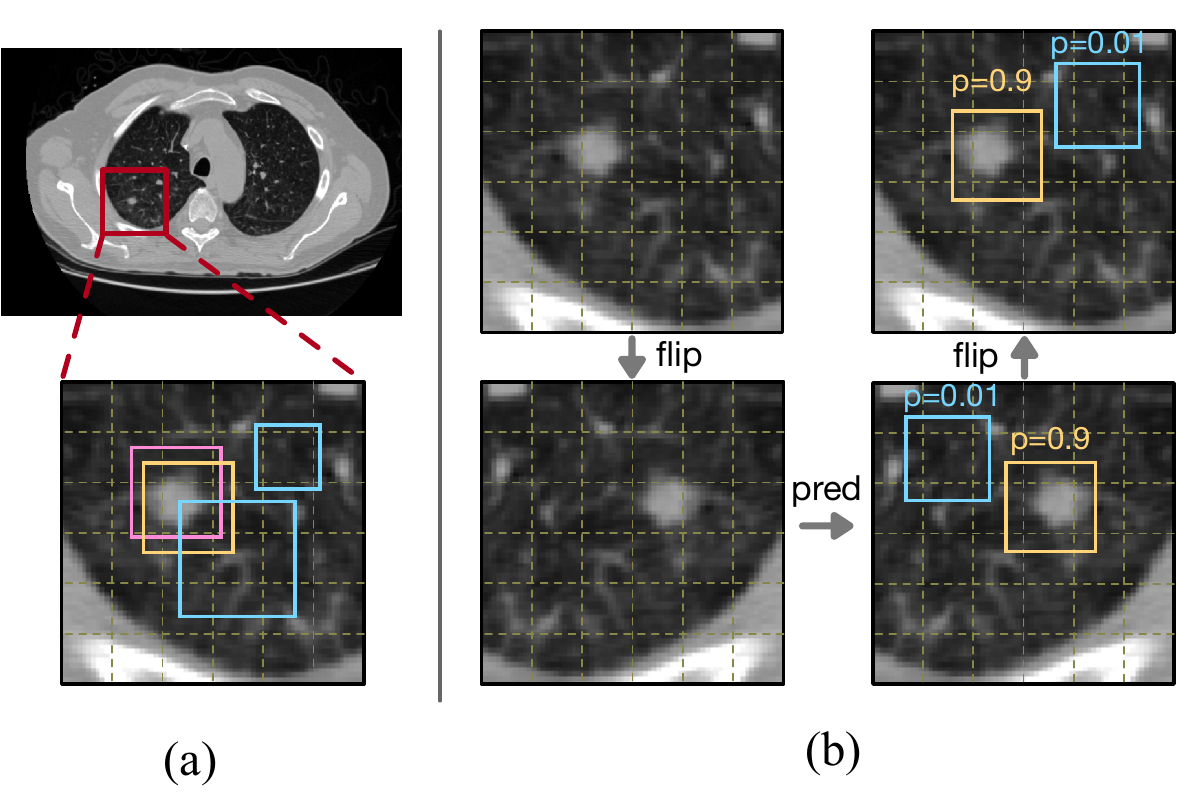}
\caption{(a) is an example of assigning targets to anchors. The dashed grids represent output feature maps where anchor boxes are defined, and each bin in the grids corresponds to a point in the feature map.
The pink box is a ground-truth bounding box. The orange box is a positive anchor and the blue boxes are negative anchors. (b) is an example of our augmentation method used for target prediction. We use flip augmentation for the image patch and predict the probability for each anchor with the model.
After that, an inverse transformation is applied to the patch and anchors. 
We only show two example anchors for illustration purposes and use consistent coloring for each anchor. Note that anchors in 3D images are also three-dimensional, of which we only show 2D slices for better visualization.
}
\label{fig: anchor}
\setlength{\abovecaptionskip}{0pt}
\setlength{\belowcaptionskip}{1pt}
\end{figure}

Anchor boxes are predefined bounding boxes densely tiling on images to match targeted objects. Following \cite{ren2015faster}, anchor boxes are set to have different scales and aspect ratios in order to capture objects of different shapes. Each anchor corresponds to a pixel in the output feature map from the detector and shares the same center with its receptive field. 
Mini-networks implemented by convolutional layers are used to make prediction for each anchor in a sliding-window manner.
During training, an anchor box is regarded as a positive anchor that matches an object if and only if it is highly overlapping with a certain ground-truth bounding box in terms of intersection over union (IoU).
Figure \ref{fig: anchor}(a) shows an example. 
During inference, the network predicts an objectness score (a.k.a. confidence score) and coordinate offsets for each anchor box as output. 
Feature Pyramid Network~\cite{lin2017feature} puts anchors on multi-scale feature maps to enhance the detection performance of small objects.

\subsubsection{Focal Loss}

The anchor assignment method leads to very few positive anchors relative to negative ones, which is called the foreground-background imbalance by Lin \etal~\cite{lin2017focal}. 
To mitigate this problem, they introduce the focal loss:
\begin{equation}
\label{eq: focal}
\setlength{\abovedisplayskip}{2pt}
\setlength{\belowdisplayskip}{2pt}
FL(p_t)=-\alpha_t(1-p_t)^\gamma \log(p_t)
\end{equation}
\begin{equation}
\setlength{\abovedisplayskip}{2pt}
\setlength{\belowdisplayskip}{2pt}
p_t=
\begin{cases}
p& \text{if $y=1$}\\
1-p& \text{otherwise}.
\end{cases}
\end{equation}
where $y\in \{0, 1\}$ is the ground-truth label for an anchor, $p$ is the model's estimated probability of the anchor being a positive example, while $\alpha_t$ is a weighting factor for different classes (namely, $\alpha_0$ and $\alpha_1$ for class 0 and 1, respectively) to balance the importance for positive and negative examples, $\gamma$ is the focusing parameter. 
The meaning of $p_t$ can be considered as the prediction confidence so that the second term in Eq. (\ref{eq: focal}) is used to down-weight confident examples to make the model focus on hard (less confident) ones.

\subsection{Semi-supervised Learning}
\label{sect: ssl}
Semi-supervised learning (SSL) aims to make use of unlabeled data to improve model performances. In this section, we briefly review an SSL framework called MixMatch~\cite{DBLP:journals/corr/abs-1905-02249}, on which our work is mainly built. MixMatch is not only one of the state-of-the-art SSL approaches, but also a unified framework that integrates the spirits of most successful attempts in this line of research (e.g., entropy minimization~\cite{grandvalet2005semi}, consistency regularization~\cite{sajjadi2016regularization} and MixUp augmentation~\cite{DBLP:conf/iclr/ZhangCDL18}). The central thesis of this work is to take MixMatch as a typical example to show how, if feasible, to tailor a general SSL approach for the medical imaging domain. In other words, our contribution is mostly orthogonal to the progress being made in SSL.

MixMatch consists of two major components, target prediction for unlabeled data and MixUp augmentation.  The first component requires to define a set of stochastic transformations of a given datapoint (e.g., an image) in such a way that its semantics (e.g., class label) barely change. In the example of image classification, rotating and shearing are two widely-used augmentations. MixMatch uses the average ensemble of predictions by the current model parameterized by $\theta$ on $K$ augmented instances $\hat{u}_{k}$ of each unlabeled training sample $u$ as "guesses" for their labels, formally,
\begin{equation}
\setlength{\abovedisplayskip}{4pt}
\setlength{\belowdisplayskip}{4pt}
\bar{y} = \frac{1}{K}\sum_{k = 1}^K \text{p}_\text{Model}(\hat{u}_{k};\theta).
\end{equation}
Then, these guessed labels are further transformed by a sharpening operator before used as training targets. The sharpening operator (for the i-th of $L$ classes) is defined by
\begin{equation}
\setlength{\abovedisplayskip}{4pt}
\setlength{\belowdisplayskip}{4pt}
    \text{Sharpen}(\bar{y}, T)_i = \bar{y}_i^{\frac{1}{T}}\bigg/ \sum_{j = 1}^L \bar{y}_j^{\frac{1}{T}},
    \label{eq:sharpen}
\end{equation}
where $T$, termed as temperature, controls the smoothness of the output distribution (as $T\to 0$, the output becomes a one-hot vector).
The sharpening operation implicitly enforces the model to output low-entropy predictions on unlabeled data. 
Once training targets for unlabeled data are available, MixMatch further utilizes the MixUp augmentation~\cite{DBLP:conf/iclr/ZhangCDL18} for both labeled and unlabeled data. More specifically, given a labeled (or unlabeled) data point with its label (or predicted target) namely $(x, y)$, MixUp augmentation produces a stochastic linear interpolation with another training example $(x', y')$, either labeled or unlabeled, as follows
\begin{align}
   \lambda &\sim \text{Beta}(\eta, \eta), \label{eq: beta}\\
   \tilde{\lambda} &= \max(\lambda, 1 - \lambda),\\
   \hat x &= \tilde{\lambda} x + (1 - \tilde{\lambda})x', \\
   \hat y &= \tilde{\lambda} y + (1 - \tilde{\lambda})y'.\label{eq: paug}
\end{align}

After the above procedures, we can get a collection of augmented training examples with supervision signals from both labeled and unlabeled data, and then use the supervised objectives to train model parameters. 


\begin{figure*}
\centering
\includegraphics[width=0.9\textwidth]{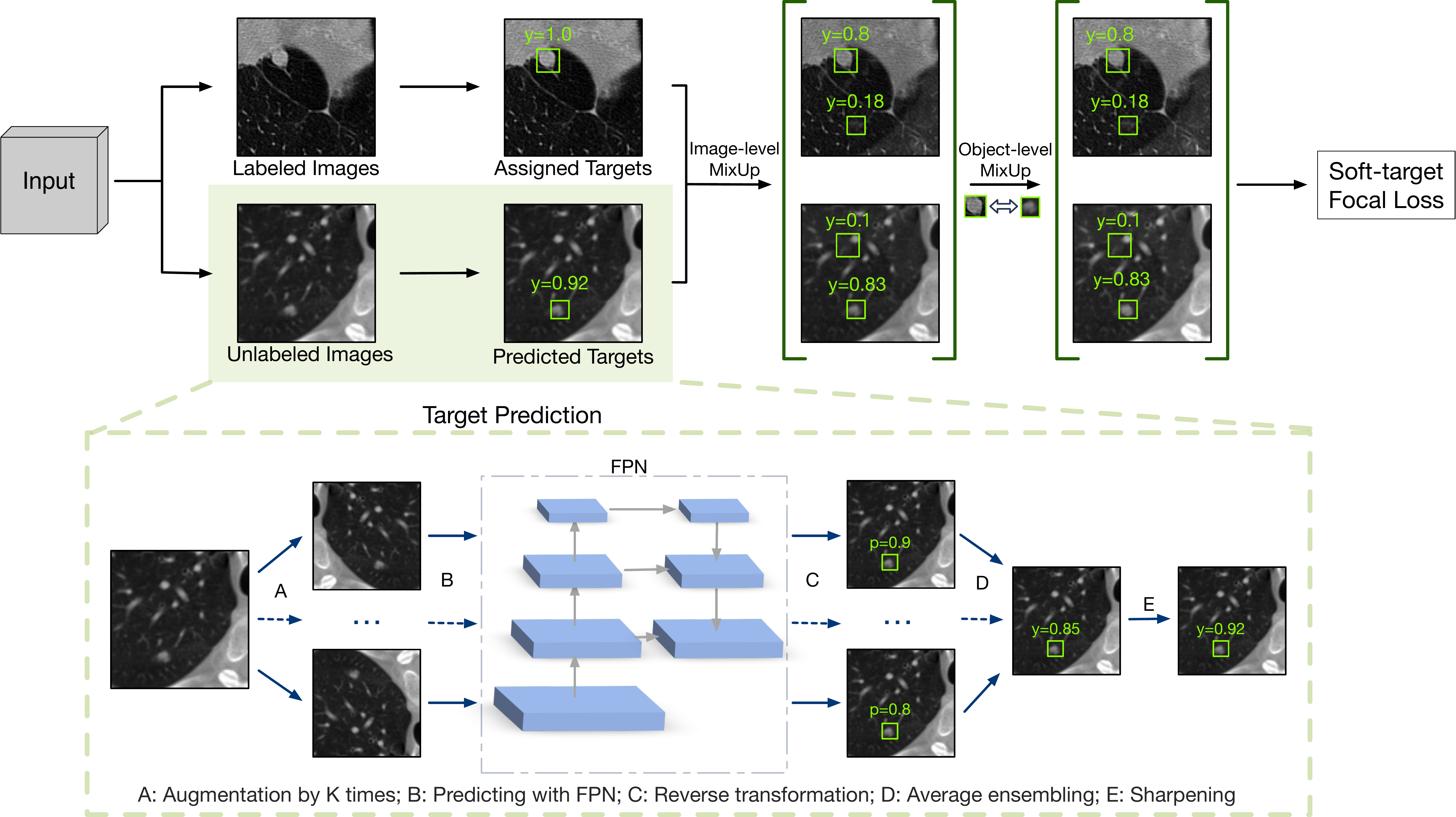}
\caption{\bd{Overview of our proposed method FocalMix.} For an input batch, the training targets of anchors in labeled images are assigned according to annotated boxes, while the unlabeled are predicted with the current model as shown in the lower part of the figure. After applying two levels of MixUp to the entire batch, we use the proposed soft-target focal loss to train the model. Throughout this paper, we only show a slice of each 3D CT scan with 3D anchors on it for ease of presentation.}
\vspace{-0.3cm}
\label{fig:framework}
\end{figure*}

\section{Methodology}
In this paper, we attempt to leverage modern semi-supervised learning methods in medical image detection. To achieve such goal, two essential components in the MixMatch framework introduced in Sect.~\ref{sect: ssl} are tailored specifically for lesion detection tasks: \textit{target prediction} and \textit{MixUp augmentation}. 
Before that, we first propose a generic generalization of the focal loss, which allows us to train detection models with \textit{soft} training targets that occurred in most modern SSL frameworks. The overview of our proposed method is shown in Figure \ref{fig:framework}.

\subsection{Soft-target Focal Loss}
\label{sect: soft focal}
Semi-supervised learning often involves soft training targets (e.g., $\hat y$ in Eq. (\ref{eq: paug})). 
This has rarely been raised as an issue in SSL literature because most current work focuses on classification tasks, and the cross-entropy loss used in classification can naturally deal with soft labels.
However, as introduced in Sect.~\ref{sect: retinanet}, the state-of-the-art object detection approaches generally use the focal loss that adds two weighting terms to the original cross-entropy loss, i.e., $\alpha(y)$ and $\beta(y, p)=(1-p_t)^\gamma$ in Eq. (\ref{eq: focal}). 
Both of the two terms are dependent on class labels, which is emphasized by writing them as functions of $y$, and, unfortunately, have no trivial extension if $y$ can take any continuous value between 0 and 1.
This is one of the major factors that hinder us from directly utilizing the off-the-shelf SSL methods.
Therefore, our proposed approach generalizes these two terms to the case of soft targets accordingly.

The first term is originally designed for class imbalance and usually proportional to the inverse frequency of class $y$. More specifically, $\alpha$ for the less frequent positive examples is larger than that for negative examples to prevent the latter from dominating the total loss.
In our case, this problem amounts to having a skewed distribution of soft labels. Hence, $\alpha(y)$ should preferably be inversely proportional to the probability density function of $y$. However, it is not very computationally feasible to do density estimation along the way of model training.
Thus, we assume that the density function of $y$ decays in roughly the same rate as $1/y$, and the density at 0 and 1 are treated as hyper-parameters to be determined by cross-validation denoted by $\alpha_0$ and $\alpha_1$, respectively. Under this assumption, we can derive the form of $\alpha(y)$ for soft labels as $\alpha(y) = \alpha_0 + y (\alpha_1 - \alpha_0)$. 

The second term $(1-p_t)^\gamma$ is used to down-weight easy examples (esp., background anchors) that are pervasive in the training process of detection models. We can interpret this term as the discrepancy between prediction ``confidence score'' $p_t$ and its target value (i.e., 1), by which the hardness of training examples can be measured to some extent. From this perspective, we can naturally generalize this term to soft-target labels by rewriting it as the $\gamma$-th power of the absolute difference between model prediction $p$ and its training target $y$, i.e., $\beta(y, p) = \vert y-p \vert^\gamma$.

To sum up, the proposed \textit{soft-target focal loss} for SSL is 
\begin{equation}
    SFL(p) = [\alpha_0 + y (\alpha_1 - \alpha_0)]\cdot \vert y-p \vert^\gamma \cdot CE
    (y, p),
\end{equation}
where $CE(y,p)=-y \log p - (1-y) \log (1-p)$ denotes the cross-entropy loss. 
We can check that \textit{focal loss} is a special case of our proposed soft-target focal loss when $y \in \left\{ 0, 1 \right\}$.

\subsection{Anchor-level Target Prediction}
\label{sec:anchorprediction}
Target prediction for unlabeled data is a widely used component in both traditional and modern approaches for SSL. However, how we can transfer existing target prediction methods from classification to detection is not a trivial question, because detection models output bounding boxes for targeted objects as opposed to more structured class labels. In FocalMix, we propose to approach this problem at the anchor level.

Following the common practice in computer vision, we sample patches of the same size ($160 \times 160 \times 160$ in our experiments) from original images during training. We also ensure that the edge length of image patches (e.g., 160) is divisible by the maximum strides (e.g., 16) used in FPN. Consequently, each anchor in an image patch can always fall into the position of another anchor after rotating or flipping.
We define the augmentation for each patch as applying these two types of transformations on it. It is worthwhile mentioning that there are richer combinations of rotation and flipping in different directions for 3D medical images than those in the 2D case (48 different combinations versus eight).  Then, we use the model to predict the probability of each anchor matching an object in the transformed image patch. After that, we can obtain a guessed target for each anchor in the original patch by an inverse transformation (rotating or flipping backward). The reader can find an intuitive example in Figure~\ref{fig: anchor}.

As shown in Figure~\ref{fig:framework}, we repeat the data augmentation procedure described above $K$ times and generate $K$ guessed targets for each anchor in a patch. Then, we aggregate these guessed targets for every anchor by the average ensemble. Finally, we apply an anchor-wise sharpening operation as in Eq. (\ref{eq:sharpen}) to obtain a low-entropy predicted target for a given patch to be used in model training.

\subsection{MixUp Augmentation for Detection}
\label{sect: mixup for detection}
MixUp augmentation is an important component in the MixMatch framework, which encourages the model to behave linearly in-between training examples for better generalization performance. The vanilla MixUp procedure is designed for image classification settings where each image is associated with one class label, while medical images are annotated with bounding boxes for diagnosed lesions in our task. 
Thus, the vanilla MixUp augmentation cannot be utilized directly. In this paper, we introduce two adapted MixUp approaches for lesion detection in medical images: image-level MixUp and object-level MixUp (see Figure~\ref{fig:mixup} for illustrative examples). 

\textbf{Image-level MixUp}. The difficulty mainly lies in how to mixup training targets as we mixup two images. Although the actual labels for detection tasks in medical imaging are bounding boxes, we cannot get something as meaningful as soft classes in classification by taking the linear interpolation of two sets of boxes.  Instead, we propose to mixup training signals at the anchor level. Formally, given two medical images of the same size along with their training targets (either annotated labels or predicted targets) for each anchor, $(x, \{y_i\})$ and $(x', \{y'_i\})$, we generate an augmented sample $(\hat x, \{ \hat y_i\})$ as follows.
\begin{align}
   \lambda &\sim \text{Beta}(\eta, \eta), \label{eq: beta_}\\
   \tilde{\lambda} &= \max(\lambda, 1 - \lambda),\\
   \hat x &= \tilde{\lambda} x + (1 - \tilde{\lambda})x', \label{eq: x_}\\
   \hat y_i &= \tilde{\lambda} y_i + (1 - \tilde{\lambda})y_i', \forall i.
\end{align}

The image-level MixUp has a more intuitive interpretation in lesion detection tasks, the goal of which is to discriminate lesion from background textures. Anchor-to-anchor mixup requires the model to be able to detect lesions that are mixed with stronger background noises than usual, analogous to the idea of ``altitude training''.

\textbf{Object-level MixUp}. 
In medical imaging tasks, objects (i.e., lesions) contain much more information than background textures, but the number of objects is often limited (only one lesion per medical image in most of the time). Therefore, we propose to generate extra object instances by mixing up different lesion patterns within each training batch.
In other words, for each object within each image in a training batch, we randomly sample another object from the current batch, re-scale it to the same size, and then mixup the two objects in the same manner as in Eq. (\ref{eq: beta_}-\ref{eq: x_}).
Note that objects are simply the annotated boxes for labeled images, while, for unlabeled ones, predicted boxes with high prediction confidence are treated as detected objects.
Since all of these objects have quite consistent targets (with high probabilities being a positive example), we no longer mixup training targets for simplicity.

\section{Experiments}
\label{sect: exp}
\begin{table*}[!htb]
\begin{center}
\begin{tabular}{|c|c|ccccccc|c|c|}
\hline
\mrtwo{Labeled} & \mrtwo{Unlabeled} & \multicolumn{7}{c|}{Recall(\%) @ FPs} & \mrtwo{CPM(\%)} & \mrtwo{Improv.} \\
\cline{3-9}
&& 0.125 & 0.25 & 0.5 & 1 & 2 & 4 & 8 &  & \\
\hline\hline
25  & -   & 46.7 & 54.0 & 60.6 & 68.6 & 74.4 & 79.1 & 82.4 & 66.6 & \mrtwo{\bd{11.5 (17.3\%)}}\\
25  & 400 & 57.6 & 64.5 & 74.6 & 80.5 & 87.0 & 90.1 & 92.1 & \bd{78.1} &  \\
\hline
50  & -   & 57.2 & 65.7 & 71.4 & 77.9 & 82.6 & 85.6 & 87.2 & 75.4 & \mrtwo{\bd{6.6 (8.8\%)}}\\
50  & 400 & 64.1 & 71.0 & 78.7 & 85.2 & 89.3 & 92.3 & 93.5 & \bd{82.0} &  \\
\hline
100 & -   & 64.9 & 73.8 & 79.7 & 85.2 & 89.0 & 92.3 & 94.5 & 82.8 & \mrtwo{\bd{4.4 (5.3\%)}}\\
100 & 400 & 73.4 & 80.9 & 84.8 & 88.6 & 92.3 & 94.7 & 96.1 & \bd{87.2} &  \\
\hline
\end{tabular}
\end{center}
\caption{\bd{Main results on the LUNA16 dataset.} We evaluate FocalMix with \{25, 50, 100\} labeled CT scans, respectively. \textit{Improv.} denotes the improvements in CPM over the fully-supervised baseline (relative improvements shown in parentheses).}
\label{tab:main}
\vspace{-0.2cm}
\end{table*}

We evaluate our proposed semi-supervised framework FocalMix on the pulmonary nodule detection task. Experiments are conducted on the \textbf{LUNA16} dataset, which is the most widely used one in pulmonary nodule detection literature. We also use the \textbf{NLST} dataset as an additional source of unlabeled data for further evaluation. 

\textbf{LUNA16}~\cite{luna16} is a high quality subset of the LIDC-IDRI dataset~\cite{lidc}. It consists of 888 thoracic CT scans in total, with 1186 annotated nodules larger than $3$mm. 
All the annotations are agreed by at least three (out of four) radiologists.
Other confusing nodules and non-nodules are marked as ``irrelevant findings'', which are counted as neither false positive nor true positive during evaluation.

\textbf{NLST}~\cite{nlst} (National Lung Screening Trial) was originally built to compare the effectiveness of thoracic CT and chest X-ray for detecting lung cancer. There are about 75,000 CT scans in the NLST dataset with the characteristics of participants, scanning test results, diagnostic procedures, etc. Since such annotations as nodule locations are not available in this dataset, we only use it as an extra unlabeled dataset after a selection process as described in Sect.~\ref{nlst_exp}.

\textbf{Evaluation.} Following \cite{luna16}, we use Free-Response Receiver Operating Characteristic (FROC) and Competition Performance Metric (CPM) to measure detection performance. The overall score of CPM is defined as the average recalls when false positive rates are 1/8, 1/4, 1/2, 1, 2, 4, and 8 FPs per scan. Although some relevant literature uses 10-fold cross-validation on the LUNA16 dataset to calculate evaluation metrics, it is not very convenient in the semi-supervised setting where the numbers of labeled and unlabeled data might constantly change over different experiments.
Instead, we resplit this dataset into 533 CT scans for training (60\%) and 355 for testing (40\%). 
All the labeled data and unlabeled data used in semi-supervised learning are sampled from the training set in our experiments.
\subsection{Experimental setup}

\bd{Detection Model.}
Following the recommendations in \cite{oliver2018realistic}, we use exactly the same model, a 3D variant of FPN~\cite{lin2017feature}, as both the fully-supervised baseline and the base model for FocalMix. 
Since the codes used in prior work (e.g., \cite{liu20193dfpn}) are currently not available, we use our in-house implementation throughout the experiments.
In our implementation, the backbone network is a modified 3D residual network~\cite{resnet} with 20 basic residual blocks. The 3D FPN outputs four levels of features with stride \{2, 4, 8, 16\} pixels with respect to the input image, and the base anchor sizes are set to be \{4, 8, 16, 32\}, respectively. During training, we first resize the input volume to spacing$=1mm$, and then randomly crop a 3D patch of size $160 \times 160 \times 160$ as the input of 3D FPN. For fully-supervised training, we use the focal loss for objectness classification and smooth L1 loss for 3D bounding box regression as in \cite{lin2017focal}. We set $\alpha_0=0.05$, $\alpha_1=0.95$ and $\gamma=2.0$ for the focal loss. 
Anchors that have an IoU with ground-truth box higher than 0.3 and smaller than 0.1 are set to be positive and negative, respectively, while others are neglected during training. 
The model is trained end-to-end using the ADAM optimizer~\cite{DBLP:journals/corr/KingmaB14} with batch size 8. We start the learning rate from 0.001 and use cosine annealing strategy~\cite{coslr}. If not specified otherwise, we train the model for 800 epochs.

\bd{Semi-supervised Learning.}
In the SSL setting, we use the same amount (more specifically, eight) of labeled data and unlabeled data in a batch of input. We apply soft-target focal loss on unlabeled data. 
We set $\alpha_0=0.05$ and $\alpha_1=0.95$ in order to be consistent with those in the supervised loss. The other settings remain the same as in the supervised version. For MixUp augmentation, image-level MixUp is first applied and then followed by object-level MixUp. We use $\eta=0.2$ for MixUp and $T=0.7$ for sharpening  throughout the experiments.

\bd{Fully-Supervised Baseline Performance.}
As suggested in \cite{oliver2018realistic}, newly proposed SSL frameworks should be \textit{compared with} and also \textit{built upon} well-tuned strong fully-supervised baselines for fair evaluation. 
Therefore, before presenting the main results in our experiments, we first compare the performance of our base model (i.e.,  an in-house implementation of 3D FPN) with the state-of-the-art results reported by other researchers on this dataset by using exactly the same 10-fold cross-validation protocol.
The results are shown in Table \ref{tab:baseline}.
Since we only focus on detection model itself, post-processing methods, such as lung segmentation to reduce false positives, are not used in our implementation, which can further improve the CPM scores. 
We can conclude from the table that our base model can achieve a comparable performance to various strong state-of-the-art single-stage detection methods.
We also report its performance on our own data spilt, which is used as the fully-supervised baseline in the experiment with an additional external source of unlabeled data (Sect.~\ref{nlst_exp}).

\begin{table}[!htb]
\begin{center}
\scalebox{1.0}{
\begin{tabular}{l|c|p{0.5in}<{\centering}}
\hline
Method & Data Split & CPM(\%)  \\
\hline\hline
DeepLung \cite{zhu2018deeplung} & 10-fold & 84.2\\
DeepSeed \cite{DBLP:journals/corr/abs-1904-03501} & 10-fold & 86.2\\
S4ND \cite{khosravan2018s4nd} & 10-fold & 89.7 \\
3D FPN \cite{liu20193dfpn} & 10-fold & 91.9 \\
Our base model & 10-fold & 91.2 \\
\hline
Our base model & 533/355 & 89.2 \\
\hline
\end{tabular}
}
\end{center}
\caption{\bd{Performance of the base model used in our experiments.} Our re-implemented 3D FPN is comparable with state-of-the-art single-stage nodule detection models.}
\label{tab:baseline}
\vspace{-0.3cm}
\end{table}

\begin{figure}
\centering
\includegraphics[width=2.5in]{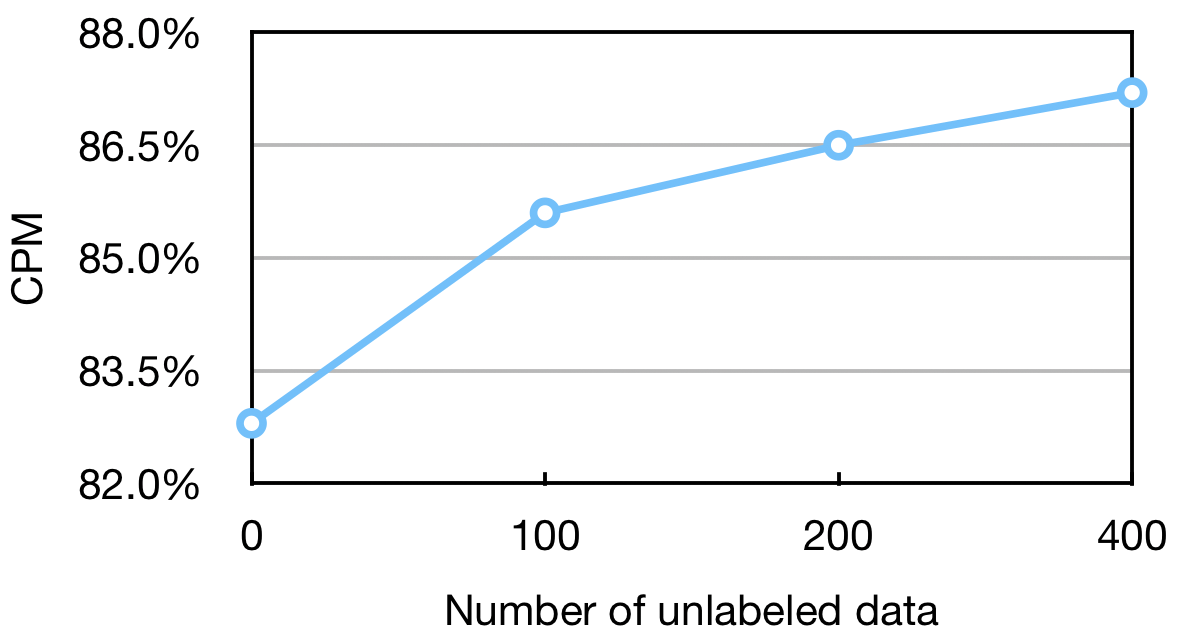}
\caption{\bd{Performance with different amounts of unlabeled data on LUNA16.} We use 100 labeled images.}
\label{fig:unlabeled}
\vspace{-0.4cm}
\end{figure}


\begin{table*}  
\centering
\subtable[Loss function.]{\scalebox{0.833}{

    \begin{tabular}{l|c}
    \hline
    Loss Function & CPM(\%) \\
    \hline
    Supervised & 82.8 \\
    SFL w/o soft $\alpha$, $\beta$ & Fail \\
    SFL w/o soft $\alpha$ & 84.4 \\
    SFL w/o soft $\beta$ & 83.7 \\
    SFL & \bd{85.2} \\
    \hline
    \end{tabular}
}}
\qquad  
\subtable[Augmentation times (K).]{          
    \begin{tabular}{p{0.4in}<{\centering}|p{0.5in}<{\centering}}
    \hline
    K & CPM(\%) \\
    \hline
    1 & 85.9 \\
    2 & 86.3 \\
    4 & \bd{87.2} \\
    8 & 87.1 \\
    \hline
    \end{tabular}
}  
\qquad  
\subtable[MixUp method.]{          
    \begin{tabular}{cc|p{0.4in}<{\centering}}
    \hline
    \multicolumn{2}{c|}{MixUp Level} & \mrtwo{CPM(\%)} \\
    \cline{1-2}
    Image & Object & \\
    \hline
    - & - & 85.2 \\
    \checkmark & - & 86.7 \\
    \checkmark &\checkmark & \bd{87.2} \\
    \hline
    \end{tabular}
}
\caption{\bd{Ablation study.} Models are trained with 100 labeled scans and 400 unlabeled ones. \textit{Fail} denotes a divergent result.}
\label{tab:ablation}
\vspace{-0.3cm}
\end{table*} 

\subsection{Main Results}
\label{sect: main results}
Table \ref{tab:main} shows the performances of FocalMix on the LUNA16 dataset with different amounts of labeled data. Recalls at seven false positive rates along with the overall CPM score are reported.
Note that, for a fair comparison, we use the same subset of labeled data for a fixed amount of labeled data and the same set of unlabeled data for all cases, both sampled from the training set.
We can conclude that FocalMix can consistently outperform the fully-supervised baseline with 25, 50 and 100 annotated CT images as labeled data, respectively, by leveraging 400 unlabeled raw images.
When we have 25 labeled images, the fully-supervised model can only obtain a CPM score of 66.6\%, whereas FocalMix boosts it to 78.1\% with a 17.3\% relative improvement. On the other hand, with 100 labeled data, even though the fully-supervised model already achieves a CPM of 82.8\%, FocalMix can still substantially enhance its performance by a 4.4\% absolute improvement.

We can also observe from Table \ref{tab:main} that, by utilizing 400 unlabeled CT scans, FocalMix can achieve a comparable result with the fully-supervised baseline that uses twice the amount of labeled data.
In other words, merely collecting 400 raw CT scans from databases has roughly the same effect of having 50 carefully annotated ones.
Furthermore, it is interesting to see that our proposed SSL approach FocalMix can get a reasonably close CPM score (87.2\%) with 100 labeled as well as 400 unlabeled scans to the fully-supervised learning result (89.2\%) with 533 labeled scans.

Figure~\ref{fig:unlabeled} shows the performance with varying numbers of unlabeled CT scans. We can observe that, the CPM score consistently grows as the amount of unlabeled data increases, which proves the effectiveness of using unlabeled data in FocalMix.

\subsection{Ablation Study}
\label{sect: ablation}
In this section, we investigate the effectiveness of different components (viz., loss function, target prediction method, MixUp augmentation strategy) in our proposed semi-supervised approach through ablation studies on the LUNA16 dataset. Since too little labeled training data can lead to unstable results, we use 100 labeled images for all the following experiments.

\bd{Loss Function.}
Our proposed soft-target focal loss generalizes the focal loss by adapting each of its term to accommodate soft targets. 
Since the cross-entropy loss can naturally deal with soft labels, only the first two terms, namely $\alpha(y)$ and $\beta(y, p)$, are modified.
To study the contributions of our extension to each of these two terms respectively, we compare the proposed loss with its degenerated version by using ``pseudo-hard targets''.
That is, we regard soft targets with probability greater than 0.5 as positive examples, and the others as negative examples. In this way, we can use the $\alpha$ and $\beta$ terms in the original focal loss in our SSL framework. As shown in Table~\ref{tab:ablation}(a), we can see that it hurts the detection performance by using either $\alpha$ or $\beta$ in their degenerated version with pseudo-hard targets (even diverged when excluding both), which demonstrates the contribution of our designed soft-target generalization to the focal loss.

\bd{Targets Prediction.}
During the target prediction stage, we first make predictions on $K$ different augmentations and ensemble the predictions by taking averaging at the anchor-level. To demonstrate the contribution of this ensemble process, we report the CPM scores of FocalMix over different $K$ in Table \ref{tab:ablation}(b). We see that it can only get a CPM score of 85.9\% when using a single augmentation for target prediction, while the CPM score improves by 1.3\% as the number of augmentations $K$ increases to 4, which validates the effectiveness of our ensemble strategy. However, we can also notice that the performance starts to saturate when $K=4$. Thus, we choose $K=4$ throughout the experiments.
 
\begin{figure*}
\centering
\includegraphics[width=0.9\textwidth]{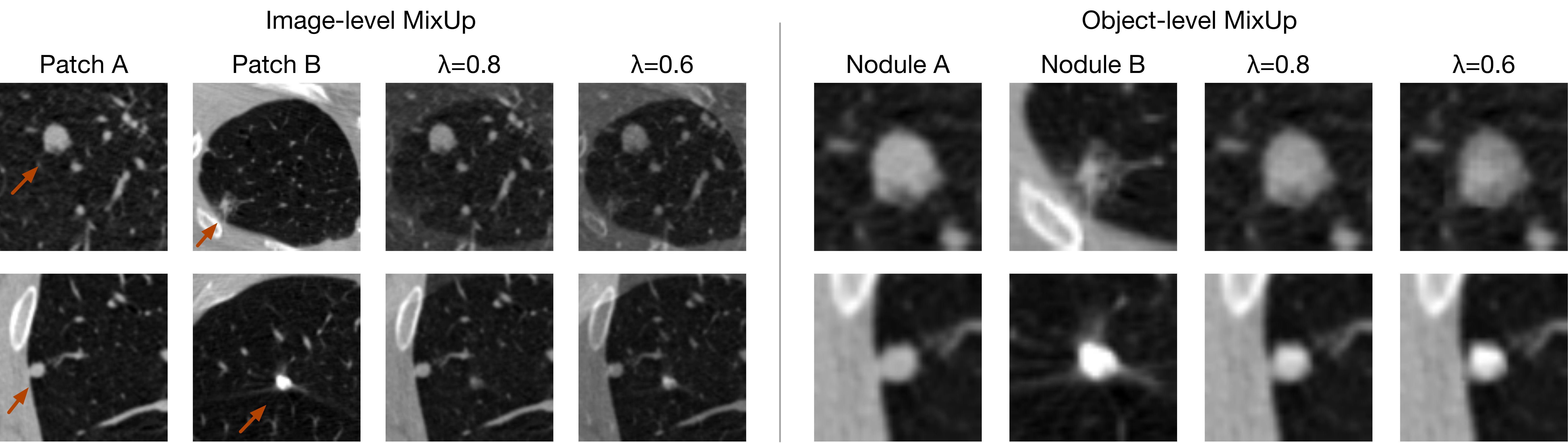}
\caption{\bd{Illustrative examples for two MixUp methods.} The left figure shows the image-level MixUp, where red arrows point to nodules in the original image. The right figure demonstrates the object-level MixUp, where we zoom in on the nodules and locate them in the center of each image patch for better visualization.}
\label{fig:mixup}
\end{figure*}

\bd{MixUp Augmentation.}
In FocalMix, two MixUp strategies are designed for medical images: image-level MixUp and object-level MixUp. As shown in table \ref{tab:ablation}, the image-level MixUp can boost the CPM score from 0.852 to 0.867, and the object-level MixUp further improves the result to 0.872. We also illustrate some examples of MixUp in Figure \ref{fig:mixup}. 
Intuitively, the goal of image-level MixUp is to encourage models to perform linearly between foreground and background, while object-level MixUp encourages models to detect lesions with richer patterns.

\subsection{SSL with More Labeled and Unlabeled Data}
\label{nlst_exp}

In previous sections, we analyze the performance of FocalMix with relatively small amounts of labeled data. 
Although this is arguably the most common scenario in real-world applications, it is also an interesting research question whether SSL can still boost the performance of supervised learning when a large training set is available. In addition, there is usually a mismatch between data distributions of labeled data and unlabeled data.
Therefore, we also evaluate our proposed SSL framework by using all the 533 CT scans from LUNA16 as labeled data and using an external database NLST (with potentially different data distribution to that of LUNA16) to sample unlabeled data.

\bd{Data Selection.} 
The NLST dataset contains $\sim$75,000 CT scans, a large number of which do not contain nodule findings.
Thus, we attempt to filter out these irrelevant images without nodules. Specifically, we first train a 3D FPN using LUNA16, make predictions on a random subset of NLST, and then pick out the CT scans that have at least one predicted nodule with high confidence (the threshold for positive nodules is set to 0.8). After selection, we leave $\sim$3,000 scans as unlabeled training data.

\bd{Results.}
The results are shown in Table \ref{tab:nlst}. We train all the models for 400 epochs. 
When using all the 533 annotated CT scans, our proposed MixUp strategies (i.e., anchor-level and object-level MixUp) alone can improve the CPM score of the fully-supervised learning approach from 89.2\% to 90.0\%.
FocalMix further improves this result to 90.7\% by leveraging around 3,000 images without annotation.

\begin{table}
\begin{center}
\begin{tabular}{l|p{0.5in}<{\centering}}
\hline
Model & CPM(\%) \\
\hline
Fully-supervised & 89.2  \\
Fully-supervised w/ MixUp & 90.0  \\
FocalMix & \bd{90.7}  \\
\hline
\end{tabular}
\end{center}
\caption{\bd{FocalMix with larger scale labeled and unlabeled data.} We use all the labeled data in LUNA16 and unlabeled data selected from NLST.}
\label{tab:nlst}
\vspace{-0.5 cm}
\end{table}

\section{Related Work}
\bd{Detection in 3D Medical Images.}
Due to limited space, we primarily review lung nodule detection methods, which is the most mature field in 3D medical images due to publicly available datasets.
Earlier lung nodule detectors use machine learning techniques with hand-craft features such as spherical filter~\cite{van2015novel, yanagihara2015pulmonary, chen2012automatic, akram2016pulmonary}. Recent prosperity of deep learning brings the success of modern object detection to the area of medical image detection. Ding \etal~\cite{ding2017accurate} propose to use 2D Faster R-CNN and 3D CNN for more accurate nodule detection. Another line of research \cite{liao2019evaluate, zhu2018deeplung, khosravan2018s4nd, DBLP:journals/corr/abs-1904-03501} uses 3D region proposal networks~\cite{ren2015faster} or feature pyramid network~\cite{lin2017feature} to detect nodule directly. Pezenshk \etal~\cite{pezeshk20183d} and Liu \etal~\cite{liu20193dfpn} further propose to use another network followed by 3D FPN to reduce false positives.

\bd{Semi-Supervised Learning.}
Most of recent studies focus on how to apply a loss term onto the unlabeled data for better generalization. Pseudo-label~\cite{lee2013pseudo} uses the predicted classes with the highest confidence as the training labels for unlabeled data. $\Pi$-Model~\cite{DBLP:conf/iclr/LaineA17} and $\Gamma$-Model~\cite{rasmus2015semi} use consistency regularization terms to penalize inconsistent predictions. Tarvainen and Valpola~\cite{tarvainen2017mean} propose to regularize models with a ``mean teacher'' using moving average of historical parameters. MixMatch~\cite{DBLP:journals/corr/abs-1905-02249} and UDA~\cite{DBLP:journals/corr/abs-1904-12848} integrate consistency regularization and modern data augmentation techniques into a unified framework, achieving a large improvement. There are also some works focusing on improving detection model by using extra images with image-level annotations~\cite{tang2016large, gao2019note}.

\bd{SSL in Medical Image Processing.}
Due to the difficulty of data annotation, SSL is widely used in medical imaging processing. Su \etal~\cite{su2019local} propose a semi-supervised nuclei classification method by using local and global consistency regularization. Ganaye \etal~\cite{ganaye2018semi} and Chen \etal~\cite{chen2019multi} also propose SSL approaches to get better segmentation results in brain images. Zhou \etal~\cite{zhou2019collaborative} improve the performance of disease grading and lesion segmentation by semi-supervised learning. ASDNet~\cite{nie2018asdnet} uses an attention-based semi-supervised learning method to boost the performance of medical image segmentation. These previous works are also limited to classification and segmentation, while this paper focuses on a more important and more complicated task in medical imaging, lesion detection.

\section{Conclusion}
This paper discusses a novel semi-supervised learning framework, FocalMix, which utilizes raw medical images without annotation to boost the performance of supervised lesion detection models. Extensive experiments show that FocalMix can substantially improve the performance of fully-supervised learning baselines. Our work demonstrates the feasibility of leveraging modern SSL approaches in 3D medical detection tasks.

\section{Acknowledgement}
This work is supported by National Key R\&D Program of China (2018YFB1402600), BJNSF (L172037) and Beijing Acedemy of Artificial Intelligence.

\clearpage
{\small
\bibliographystyle{ieee_fullname}
\bibliography{egbib}

\begin{thebibliography}{10}\itemsep=-1pt

\bibitem{akram2016pulmonary}
Sheeraz Akram, Muhammad~Younus Javed, M~Usman Akram, Usman Qamar, and Ali
  Hassan.
\newblock Pulmonary nodules detection and classification using hybrid features
  from computerized tomographic images.
\newblock {\em Journal of Medical Imaging and Health Informatics},
  6(1):252--259, 2016.

\bibitem{lidc}
Samuel~G Armato~III, Geoffrey McLennan, Luc Bidaut, Michael~F McNitt-Gray,
  Charles~R Meyer, Anthony~P Reeves, Binsheng Zhao, Denise~R Aberle, Claudia~I
  Henschke, Eric~A Hoffman, et~al.
\newblock The lung image database consortium (lidc) and image database resource
  initiative (idri): a completed reference database of lung nodules on ct
  scans.
\newblock {\em Medical physics}, 38(2):915--931, 2011.

\bibitem{DBLP:journals/corr/abs-1905-02249}
David Berthelot, Nicholas Carlini, Ian Goodfellow, Nicolas Papernot, Avital
  Oliver, and Colin~A Raffel.
\newblock {MixMatch}: A holistic approach to semi-supervised learning.
\newblock In H. Wallach, H. Larochelle, A. Beygelzimer, F. d\textquotesingle
  Alch\'{e}-Buc, E. Fox, and R. Garnett, editors, {\em Advances in Neural
  Information Processing Systems 32}, pages 5049--5059. Curran Associates,
  Inc., 2019.

\bibitem{chen2012automatic}
Bin Chen, Takayuki Kitasaka, Hirotoshi Honma, Hirotsugu Takabatake, Masaki
  Mori, Hiroshi Natori, and Kensaku Mori.
\newblock Automatic segmentation of pulmonary blood vessels and nodules based
  on local intensity structure analysis and surface propagation in 3d chest ct
  images.
\newblock {\em International journal of computer assisted radiology and
  surgery}, 7(3):465--482, 2012.

\bibitem{chen2018augmented}
Po-Hsuan Chen, Krishna Gadepalli, Robert MacDonald, Yun Liu, Kunal Nagpal, Timo
  Kohlberger, Greg~S Corrado, Jason~D Hipp, and Martin~C Stumpe.
\newblock An augmented reality microscope for real-time automated detection of
  cancer.
\newblock In {\em Proc. Annu. Meeting American Association Cancer Research},
  2018.

\bibitem{chen2019multi}
Shuai Chen, Gerda Bortsova, Antonio Garc{\'\i}a-Uceda Ju{\'a}rez, Gijs van
  Tulder, and Marleen de Bruijne.
\newblock Multi-task attention-based semi-supervised learning for medical image
  segmentation.
\newblock In {\em International Conference on Medical Image Computing and
  Computer-Assisted Intervention}, pages 457--465. Springer, 2019.

\bibitem{ding2017accurate}
Jia Ding, Aoxue Li, Zhiqiang Hu, and Liwei Wang.
\newblock Accurate pulmonary nodule detection in computed tomography images
  using deep convolutional neural networks.
\newblock In {\em International Conference on Medical Image Computing and
  Computer-Assisted Intervention}, pages 559--567. Springer, 2017.

\bibitem{esteva2017dermatologist}
Andre Esteva, Brett Kuprel, Roberto~A Novoa, Justin Ko, Susan~M Swetter,
  Helen~M Blau, and Sebastian Thrun.
\newblock Dermatologist-level classification of skin cancer with deep neural
  networks.
\newblock {\em Nature}, 542(7639):115, 2017.

\bibitem{ganaye2018semi}
Pierre-Antoine Ganaye, Micha{\"e}l Sdika, and Hugues Benoit-Cattin.
\newblock Semi-supervised learning for segmentation under semantic constraint.
\newblock In {\em International Conference on Medical Image Computing and
  Computer-Assisted Intervention}, pages 595--602. Springer, 2018.

\bibitem{gao2019note}
Jiyang Gao, Jiang Wang, Shengyang Dai, Li-Jia Li, and Ram Nevatia.
\newblock {NOTE-RCNN}: Noise tolerant ensemble rcnn for semi-supervised object
  detection.
\newblock In {\em Proceedings of the IEEE International Conference on Computer
  Vision}, pages 9508--9517, 2019.

\bibitem{grandvalet2005semi}
Yves Grandvalet and Yoshua Bengio.
\newblock Semi-supervised learning by entropy minimization.
\newblock In {\em Advances in neural information processing systems}, pages
  529--536, 2005.

\bibitem{gulshan2016development}
Varun Gulshan, Lily Peng, Marc Coram, Martin~C Stumpe, Derek Wu, Arunachalam
  Narayanaswamy, Subhashini Venugopalan, Kasumi Widner, Tom Madams, Jorge
  Cuadros, et~al.
\newblock Development and validation of a deep learning algorithm for detection
  of diabetic retinopathy in retinal fundus photographs.
\newblock {\em Jama}, 316(22):2402--2410, 2016.

\bibitem{resnet}
Kaiming He, Xiangyu Zhang, Shaoqing Ren, and Jian Sun.
\newblock Deep residual learning for image recognition.
\newblock In {\em Proceedings of the IEEE conference on computer vision and
  pattern recognition}, pages 770--778, 2016.

\bibitem{khosravan2018s4nd}
Naji Khosravan and Ulas Bagci.
\newblock {S4ND}: Single-shot single-scale lung nodule detection.
\newblock In {\em International Conference on Medical Image Computing and
  Computer-Assisted Intervention}, pages 794--802. Springer, 2018.

\bibitem{DBLP:journals/corr/KingmaB14}
Diederik~P. Kingma and Jimmy Ba.
\newblock Adam: {A} method for stochastic optimization.
\newblock In Yoshua Bengio and Yann LeCun, editors, {\em 3rd International
  Conference on Learning Representations, {ICLR} 2015, San Diego, CA, USA, May
  7-9, 2015, Conference Track Proceedings}, 2015.

\bibitem{cifar}
Alex Krizhevsky, Geoffrey Hinton, et~al.
\newblock Learning multiple layers of features from tiny images.
\newblock Technical report, University of Toronto, 2009.

\bibitem{DBLP:conf/iclr/LaineA17}
Samuli Laine and Timo Aila.
\newblock Temporal ensembling for semi-supervised learning.
\newblock In {\em 5th International Conference on Learning Representations,
  {ICLR} 2017, Toulon, France, April 24-26, 2017, Conference Track
  Proceedings}, 2017.

\bibitem{lee2013pseudo}
Dong-Hyun Lee.
\newblock Pseudo-label: The simple and efficient semi-supervised learning
  method for deep neural networks.
\newblock In {\em Workshop on Challenges in Representation Learning, ICML},
  volume~3, page~2, 2013.

\bibitem{DBLP:journals/corr/abs-1904-03501}
Yuemeng Li, Hangfan Liu, and Yong Fan.
\newblock {DeepSEED}: {3D} squeeze-and-excitation encoder-decoder convnets for
  pulmonary nodule detection.
\newblock {\em CoRR}, abs/1904.03501, 2019.

\bibitem{liao2019evaluate}
Fangzhou Liao, Ming Liang, Zhe Li, Xiaolin Hu, and Sen Song.
\newblock Evaluate the malignancy of pulmonary nodules using the 3-d deep leaky
  noisy-or network.
\newblock {\em IEEE transactions on neural networks and learning systems},
  2019.

\bibitem{lin2017feature}
Tsung-Yi Lin, Piotr Doll{\'a}r, Ross Girshick, Kaiming He, Bharath Hariharan,
  and Serge Belongie.
\newblock Feature pyramid networks for object detection.
\newblock In {\em Proceedings of the IEEE conference on computer vision and
  pattern recognition}, pages 2117--2125, 2017.

\bibitem{lin2017focal}
Tsung-Yi Lin, Priya Goyal, Ross Girshick, Kaiming He, and Piotr Doll{\'a}r.
\newblock Focal loss for dense object detection.
\newblock In {\em Proceedings of the IEEE international conference on computer
  vision}, pages 2980--2988, 2017.

\bibitem{liu20193dfpn}
Jingya Liu, Liangliang Cao, Oguz Akin, and Yingli Tian.
\newblock {3DFPN-HS$^2$}: {3D} feature pyramid network based high sensitivity
  and specificity pulmonary nodule detection.
\newblock In {\em International Conference on Medical Image Computing and
  Computer-Assisted Intervention}, pages 513--521. Springer, 2019.

\bibitem{coslr}
Ilya Loshchilov and Frank Hutter.
\newblock {SGDR:} stochastic gradient descent with warm restarts.
\newblock In {\em 5th International Conference on Learning Representations,
  {ICLR} 2017, Toulon, France, April 24-26, 2017, Conference Track
  Proceedings}, 2017.

\bibitem{nie2018asdnet}
Dong Nie, Yaozong Gao, Li Wang, and Dinggang Shen.
\newblock {ASDNet}: Attention based semi-supervised deep networks for medical
  image segmentation.
\newblock In {\em International Conference on Medical Image Computing and
  Computer-Assisted Intervention}, pages 370--378. Springer, 2018.

\bibitem{oliver2018realistic}
Avital Oliver, Augustus Odena, Colin~A Raffel, Ekin~Dogus Cubuk, and Ian
  Goodfellow.
\newblock Realistic evaluation of deep semi-supervised learning algorithms.
\newblock In {\em Advances in Neural Information Processing Systems}, pages
  3235--3246, 2018.

\bibitem{pezeshk20183d}
Aria Pezeshk, Sardar Hamidian, Nicholas Petrick, and Berkman Sahiner.
\newblock {3D} convolutional neural networks for automatic detection of
  pulmonary nodules in chest ct.
\newblock {\em IEEE journal of biomedical and health informatics}, 2018.

\bibitem{rasmus2015semi}
Antti Rasmus, Mathias Berglund, Mikko Honkala, Harri Valpola, and Tapani Raiko.
\newblock Semi-supervised learning with ladder networks.
\newblock In {\em Advances in neural information processing systems}, pages
  3546--3554, 2015.

\bibitem{ren2015faster}
Shaoqing Ren, Kaiming He, Ross Girshick, and Jian Sun.
\newblock {Faster R-CNN}: Towards real-time object detection with region
  proposal networks.
\newblock In {\em Advances in neural information processing systems}, pages
  91--99, 2015.

\bibitem{sajjadi2016regularization}
Mehdi Sajjadi, Mehran Javanmardi, and Tolga Tasdizen.
\newblock Regularization with stochastic transformations and perturbations for
  deep semi-supervised learning.
\newblock In {\em Advances in Neural Information Processing Systems}, pages
  1163--1171, 2016.

\bibitem{luna16}
Arnaud Arindra~Adiyoso Setio, Alberto Traverso, Thomas De~Bel, Moira~SN Berens,
  Cas van~den Bogaard, Piergiorgio Cerello, Hao Chen, Qi Dou, Maria~Evelina
  Fantacci, Bram Geurts, et~al.
\newblock Validation, comparison, and combination of algorithms for automatic
  detection of pulmonary nodules in computed tomography images: the luna16
  challenge.
\newblock {\em Medical image analysis}, 42:1--13, 2017.

\bibitem{su2019local}
Hai Su, Xiaoshuang Shi, Jinzheng Cai, and Lin Yang.
\newblock Local and global consistency regularized mean teacher for
  semi-supervised nuclei classification.
\newblock In {\em International Conference on Medical Image Computing and
  Computer-Assisted Intervention}, pages 559--567. Springer, 2019.

\bibitem{tang2016large}
Yuxing Tang, Josiah Wang, Boyang Gao, Emmanuel Dellandr{\'e}a, Robert
  Gaizauskas, and Liming Chen.
\newblock Large scale semi-supervised object detection using visual and
  semantic knowledge transfer.
\newblock In {\em Proceedings of the IEEE Conference on Computer Vision and
  Pattern Recognition}, pages 2119--2128, 2016.

\bibitem{tarvainen2017mean}
Antti Tarvainen and Harri Valpola.
\newblock Mean teachers are better role models: Weight-averaged consistency
  targets improve semi-supervised deep learning results.
\newblock In {\em Advances in neural information processing systems}, pages
  1195--1204, 2017.

\bibitem{nlst}
National Lung Screening Trial~Research Team.
\newblock Reduced lung-cancer mortality with low-dose computed tomographic
  screening.
\newblock {\em New England Journal of Medicine}, 365(5):395--409, 2011.

\bibitem{van2015novel}
Sil van~de Leemput, Frank Dorssers, and Babak~Ehteshami Bejnordi.
\newblock A novel spherical shell filter for reducing false positives in
  automatic detection of pulmonary nodules in thoracic ct scans.
\newblock In {\em Medical Imaging 2015: Computer-Aided Diagnosis}, volume 9414,
  page 94142P. International Society for Optics and Photonics, 2015.

\bibitem{DBLP:journals/corr/abs-1904-12848}
Qizhe Xie, Zihang Dai, Eduard~H. Hovy, Minh{-}Thang Luong, and Quoc~V. Le.
\newblock Unsupervised data augmentation.
\newblock {\em CoRR}, abs/1904.12848, 2019.

\bibitem{yanagihara2015pulmonary}
Takanobu Yanagihara and Hotaka Takizawa.
\newblock Pulmonary nodule detection from x-ray ct images based on region shape
  analysis and appearance-based clustering.
\newblock {\em Algorithms}, 8(2):209--223, 2015.

\bibitem{DBLP:conf/iclr/ZhangCDL18}
Hongyi Zhang, Moustapha Ciss{\'{e}}, Yann~N. Dauphin, and David Lopez{-}Paz.
\newblock Mixup: Beyond empirical risk minimization.
\newblock In {\em 6th International Conference on Learning Representations,
  {ICLR} 2018, Vancouver, BC, Canada, April 30 - May 3, 2018, Conference Track
  Proceedings}. OpenReview.net, 2018.

\bibitem{zhou2019collaborative}
Yi Zhou, Xiaodong He, Lei Huang, Li Liu, Fan Zhu, Shanshan Cui, and Ling Shao.
\newblock Collaborative learning of semi-supervised segmentation and
  classification for medical images.
\newblock In {\em Proceedings of the IEEE Conference on Computer Vision and
  Pattern Recognition}, pages 2079--2088, 2019.

\bibitem{zhu2018deeplung}
Wentao Zhu, Chaochun Liu, Wei Fan, and Xiaohui Xie.
\newblock Deeplung: Deep {3D} dual path nets for automated pulmonary nodule
  detection and classification.
\newblock In {\em 2018 IEEE Winter Conference on Applications of Computer
  Vision (WACV)}, pages 673--681. IEEE, 2018.

\end{thebibliography}
}

\end{document}